\documentclass[american]{article}
\usepackage[T1]{fontenc}
\usepackage{textcomp}
\usepackage[utf8]{inputenc}
\usepackage{array}
\usepackage{refstyle}
\usepackage{booktabs}
\usepackage{multirow}
\usepackage{varwidth}
\usepackage{graphicx}
\usepackage[authoryear]{natbib}

\makeatletter

\AtBeginDocument{\providecommand\Figref[1]{\ref{Fig:#1}}}
\AtBeginDocument{\providecommand\Tabref[1]{\ref{Tab:#1}}}
\providecommand{\tabularnewline}{\\}
\newenvironment{cellvarwidth}[1][t]
    {\begin{varwidth}[#1]{\linewidth}}
    {\@finalstrut\@arstrutbox\end{varwidth}}
\RS@ifundefined{subsecref}
  {\newref{subsec}{name = \RSsectxt}}
  {}
\RS@ifundefined{thmref}
  {\def\RSthmtxt{theorem~}\newref{thm}{name = \RSthmtxt}}
  {}
\RS@ifundefined{lemref}
  {\def\RSlemtxt{lemma~}\newref{lem}{name = \RSlemtxt}}
  {}

\usepackage{arxiv}
\usepackage{csquotes}
\usepackage{amsmath}
\DeclareMathOperator*{\argmax}{argmax}

\makeatother

\usepackage{babel}
\begin{document}
\title{Advancing TDFN: Precise Fixation Point Generation Using Reconstruction
Differences}
\author{Shuguang Wang, Yuanjing Wang}
\maketitle
\begin{abstract}
\citet{wangTaskDrivenFixationNetwork2025} proposed the Task-Driven
Fixation Network (TDFN) based on the fixation mechanism, which leverages
low-resolution information along with high-resolution details near
fixation points to accomplish specific visual tasks. The model employs
reinforcement learning to generate fixation points. However, training
reinforcement learning models is challenging, particularly when aiming
to generate pixel-level accurate fixation points on high-resolution
images. This paper introduces an improved fixation point generation
method by leveraging the difference between the reconstructed image
and the input image to train the fixation point generator. This approach
directs fixation points to areas with significant differences between
the reconstructed and input images. Experimental results demonstrate
that this method achieves highly accurate fixation points, significantly
enhances the network's classification accuracy, and reduces the average
number of required fixations to achieve a predefined accuracy level.
\end{abstract}

\section{Introduction}

\begin{figure}
\centering
\includegraphics[scale=0.65]{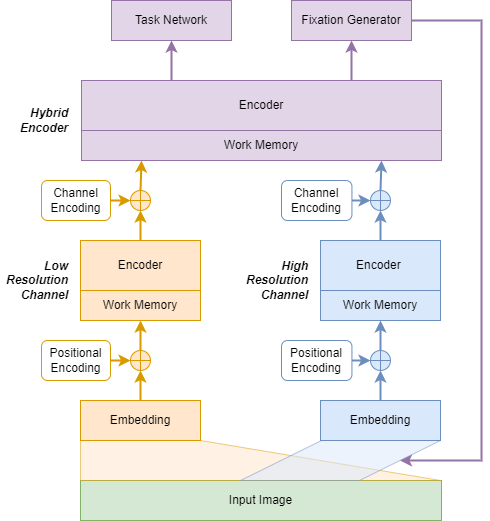}

\caption{TDFN architecture.}\label{fig:TDFN-architecture.}

\end{figure}

\citet{wangTaskDrivenFixationNetwork2025} proposed the TDFN (Task-Driven
Fixation Network), as illustrated in \Figref{TDFN-architecture.},
which accomplishes specific visual tasks through a fixation mechanism,
aiming to reduce network size and computational cost. For classification
tasks, TDFN utilizes low-resolution global image information along
with a series of high-resolution ROI inputs to perform object classification.
The selection of ROIs depends on how fixation points are generated,
making the choice of fixation points a critical factor influencing
classification performance.

They employed a reinforcement learning method to generate fixation
points. Using the MNIST dataset as an example, the original images
were down-sampled to a low-resolution size of 4\texttimes 4 pixels,
and the selection of fixation points was conducted on this low-resolution
image. However, this approach resulted in poor localization of fixation
points, failing to accurately target the critical feature areas of
the characters, thereby limiting further improvements in network performance.

Attempting to select fixation points directly on the high-resolution
original images to achieve more precise localization renders reinforcement
learning infeasible. This is primarily because reinforcement learning
relies on random sampling for optimization, but the vast number of
potential choices makes finding the optimal selection through random
trials computationally prohibitive.

Therefore, exploring more effective methods for generating fixation
points with finer localization is essential.

There is a wide range of work across different fields related to the
selection of fixation points, although various terminologies may be
used. For example, based on Feature Integration Theory \citep{treismanFeatureintegrationTheoryAttention1980},
a visual model for salient region detection was proposed \citep{kochShiftsSelectiveVisual1985,ittiModelSaliencybasedVisual1998}.
Additionally, some models have attempted to integrate bottom-up and
top-down information to determine fixation points \citep{olivaTopdownControlVisual2003,petersBottomupIncorporatingTaskdependent2007,zhangSUNBayesianFramework2008,gaoDecisionTheoreticSaliencyComputational2009,yulinxieBayesianSaliencyLow2013}.
In this context, bottom-up refers to saliency maps derived from low-level
features, while top-down encompasses target features and scene context.

In recent years, there has been a growing trend in using deep neural
networks to detect salient regions or generate fixation points \citep{vigLargeScaleOptimizationHierarchical2014,zhouLearningDeepFeatures2016,huangSALICONReducingSemantic2015,kummererDeepGazeBoosting2015,kummererDeepGazeIIReading2016,panSalGANVisualSaliency2018,wangInferringSalientObjects2020,wangPyramidVisionTransformer2021,yanReviewVisualSaliency2021}.
An interesting method proposed by \citet{simonyanDeepConvolutionalNetworks2014}
used the derivative of the classification score with respect to the
input image to obtain a class saliency map. In Transformer models,
\citet{dosovitskiyImageWorth16x162021} utilized Attention Rollout
\citep{abnarQuantifyingAttentionFlow2020} to visualize the attention
weights of output tokens on the input image, which can also be used
to localize recognition targets.

However, the requirements of TDFN are different. TDFN assumes that
the initial input consists only of a low-resolution global image,
and local high-resolution ROIs are gradually added in a sequential
manner based on fixation points. This means that the original high-resolution
input image is not accessible to TDFN, especially during the inference
stage. As a result, it is not possible to directly compute the derivative
of the classification score with respect to the input image, nor is
it feasible to perform backtracking to identify key regions. Therefore,
alternative strategies must be employed.

This paper proposes the following approach:
\begin{itemize}
\item Based on the existing input information (e.g., a low-resolution global
image or several high-resolution ROIs centered at fixation points),
the network performs internal image reconstruction to predict a high-resolution
global image.
\item Furthermore, the fixation point generator is trained to predict a
saliency map by calculating the difference between the reconstructed
image and the input image, selecting the region with the largest difference
as the next fixation point.
\end{itemize}
Experiments demonstrate that this approach is feasible and can generate
accurately localized fixation points at a high-resolution scale. On
this basis, it significantly improves the performance of TDFN, including
enhanced classification accuracy and a reduction in the average number
of required fixations to achieve a predefined accuracy level. The
reduction in the average number of fixations translates to lower computational
cost during the inference stage.

It is also worth noting that, compared to the fixed patching schema
in traditional Vision Transformer models, the TDFN model, along with
the more accurate fixation point localization method proposed in this
paper, achieves a dynamic patching mechanism with low computational
cost. Specifically, it applies finer-grained patching to salient regions
and coarser analysis to less important regions, thereby ensuring classification
accuracy.

\section{Related Works}

TDFN is based on Transformer architecture \citep{vaswaniAttentionAllYou2017}.
The computational complexity of the Transformer is $O\left(N^{2}D+ND^{2}\right)$,
where $N$ is the length of the token sequence and $D$ is the dimension
of the tokens. In ViT models \citep{dosovitskiyImageWorth16x162021},
the input image is divided into fixed-size patches (e.g., 16\texttimes 16)
and then embedded. Typically, using smaller patches improves classification
accuracy but significantly increases computational cost. For instance,
reducing the patch size from 16\texttimes 16 to 8\texttimes 8 increases
the sequence length by a factor of 4, and the self-attention computation
increases by a factor of 16. Consequently, it becomes impractical
to use finer-grained patches in ViT to improve classification performance
or for applications requiring precise localization, such as pixel-level
segmentation.

TDFN provides a flexible solution to reduce computational cost. Leveraging
the fixation mechanism, TDFN performs low-resolution analysis on the
entire image while conducting high-resolution analysis on a few key
ROI regions. In this process, selecting fixation points becomes a
critical factor. In the original work, TDFN generates fixation points
using reinforcement learning.

However, reinforcement learning struggles to produce pixel-level accurate
fixation points. Even for a relatively simple dataset like MNIST,
where the original image size is 28\texttimes 28 pixels, generating
pixel-level fixation points would mean $28\times28$ possible choices
for the first fixation point---too large for reinforcement learning
to handle effectively. To address this, the original work generates
fixation points based on a down-sampled low-resolution image (e.g.,
4\texttimes 4), reducing the number of choices and making reinforcement
learning feasible. However, the coarse localization of fixation points
negatively impacts the network's performance.

Existing approaches, such as computing the derivative of the classification
loss with respect to the input image, are also unsuitable for TDFN.
First, fixation points need to be generated during inference, where
the loss function is undefined. Second, TDFN assumes that only low-resolution
global image input and a few high-resolution ROIs are available, making
the original high-resolution global image inaccessible during inference.

This paper addresses these limitations by introducing a fixation point
generator that predicts the difference between an internally reconstructed
image and the input image. Fixation points are selected from regions
with the largest differences, enabling precise pixel-level localization
and significantly improving TDFN's efficiency and performance.

\section{Method}

\subsection{Fixation Point Selection from the Perspective of the Loss Function}

From the perspective of the loss function, the selection of fixation
points can be analyzed mathematically. Assuming the network parameters
are fixed, the loss function $L$ can be expressed as a function of
the input image $x$, i.e., $L\left(x\right)$. Let us further assume
that there exists an optimal input $x^{*}$, called the template,
such that the loss function $L\left(x^{*}\right)$ is minimized. Based
on the Taylor expansion, we have:
\[
L\left(x^{*}\right)\approx L\left(x\right)+L^{'}\left(x\right)^{T}\left(x^{*}-x\right)
\]

Assuming $L\left(x^{*}\right)$ is very small and can be ignored,
this simplifies to:
\begin{equation}
-L\left(x\right)=L^{'}\left(x\right)^{T}\left(x^{*}-x\right)
\end{equation}

From this, it can be inferred that to quickly reduce the loss, one
should select points in the input image where the derivative $L^{'}\left(x\right)$
is large and the difference between the input $x$ and the template
$x^{*}$ is significant. Therefore, selecting fixation points purely
based on the derivative of the loss function, as done in some literature,
might lead to sub-optimal results.

In TDFN, computing $L^{'}\left(x\right)$ is challenging because the
fixation mechanism only samples part of $x$ as input. Therefore,
we focus on the term $\left(x^{*}-x\right)$. By substituting the
reconstructed image for $x^{*}$, the computation of $\left(x^{*}-x\right)$
becomes significantly simpler.

\subsection{Image Reconstruction}

The visual world that we perceive is actively constructed by the brain
through complex reasoning and completion mechanisms. The raw input
from the human eye is often noisy, distorted, and unstable. Rather
than providing perfect data, the eye supplies cues for the brain to
reconstruct a coherent representation of the world.

Similarly, it is natural for a visual neural network to include an
image reconstruction module. In TDFN, the image reconstruction module
generates a reconstructed image, which can be interpreted as the system’s
expectation of the input based on prior experience. In our method,
this reconstructed image serves as the template $x^{*}$ in the earlier
formulation.

\subsection{Fixation Point Generation and Training Mechanism}

In the TDFN model shown in \Figref{TDFN-architecture.}, the output
tokens from the Hybrid Encoder (HE) are sent to both the Task Network
and the Fixation Generator. The Task Network typically consists of
a classifier and an image reconstructor. The Fixation Generator outputs
a saliency map, which is sampled to determine the next fixation point.
The process can be summarized as follows:
\begin{itemize}
\item Classification Output:
\begin{equation}
ClassScore=Classifier\left(cls\_token\right)
\end{equation}
\item Reconstruction Output: 
\begin{equation}
ReconImage=Reconstructor\left(rec\_token\right)
\end{equation}
\item Saliency Map Generation:
\begin{equation}
SaliencyMap=FixationGenerator\left(rec\_token\right)
\end{equation}
\item Next Fixation Point Selection:
\begin{equation}
NextFixationPoint=\argmax_{\left(x,y\right)}SaliencyMap
\end{equation}
\end{itemize}
Instead of the reinforcement learning approach used in the original
work, this paper trains the fixation point generator to predict the
difference between the reconstructed image and the actual input image.
The fixation generator selects the pixel with the largest difference
as the next fixation point. The loss function for the fixation generator
is defined as:
\begin{equation}
FixationLoss=MSE\left(SaliencyMap,ReconErrMap\right)
\end{equation}

where the Reconstruction Error Map is computed as:
\begin{equation}
ReconErrMap=|ReconImage-InputImage|
\end{equation}

\begin{figure}
\centering
\includegraphics{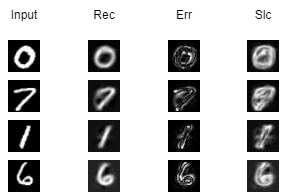}

\caption{Different maps output by TDFN. The first column shows the original
input image, the second column displays the reconstructed image, the
third column represents the absolute difference between the reconstructed
image and the input image, and the last column depicts the saliency
map generated by the fixation point generator.}\label{fig:Different-maps-output}

\end{figure}

In this framework, all functional modules (the classifier, reconstructor,
and fixation generator) can be trained simultaneously. The total loss
function is defined as:
\begin{equation}
TotalLoss=ClassLoss+\alpha\cdot ReconLoss+\beta\cdot FixationLoss
\end{equation}

where $\alpha$ and $\beta$ are adjustable coefficients. In our experiments,
we set $\alpha=0.2$ and $\beta=0.1$. The individual loss terms for
classifier and reconstructor are defined as follows:
\begin{itemize}
\item Classification Loss:
\begin{equation}
ClassLoss=CrossEntropy\left(ClassScore,labels\right)
\end{equation}
\item Reconstruction Loss:
\begin{equation}
ReconLoss=MSE\left(ReconImage,InputImage\right)
\end{equation}
\end{itemize}
Examples of the reconstructed image, reconstruction error map, and
saliency map generated by the Fixation Generator are presented in
\Figref{Different-maps-output}. The results clearly demonstrate that
the Fixation Generator produces salient points at stroke endpoints,
corners, intersections, and edges, aligning well with our expectations.

\section{Experiments}

\subsection{Dataset and Parameter Settings}

The experiments were conducted on the MNIST dataset, which consists
of 10 classes, each containing 6,000 training images and 1,000 validation
images. All images were normalized to dimensions of $32\times32$
pixels. The parameter settings are as follows:
\begin{itemize}
\item LRC Input: The original $32\times32$ pixel image was downscaled by
a factor of 8, resulting in a $4\times4$ low-resolution global image.
The patch size for embedding was set to 1 pixel.
\item HRC Input: Regions of interest (ROIs) measuring $8\times8$ pixels
were cropped from the original image. The patch size for embedding
was set to 2 pixels.
\item Embedding Dimension: The LRC and HRC modules use an embedding dimension
of 32. The HE module uses an embedding dimension of 64. To ensure
compatibility, adapters consisting of linear transformation layers
are used between the LRC, HRC, and HE modules. These adapters transform
the output tokens from the LRC and HRC modules from a dimension of
32 to a dimension of 64 for use as HE inputs.
\item Encoder Layers: Each of the LRC, HRC, and HE modules consists of 6
Transformer-based encoder layers.
\item Number of Attention Heads: All modules use 4 attention heads.
\end{itemize}

\subsection{Visualization of Fixation Points}

In previous work \citep{wangTaskDrivenFixationNetwork2025}, fixation
points were determined on low-resolution $4\times4$ images using
a reinforcement learning approach, referred to as FPG1. In this paper,
fixation points are determined on high-resolution $32\times32$ images
by predicting the differences between the reconstructed image and
the input image, referred to as FPG2.

\begin{figure}
\centering
\includegraphics[scale=0.65]{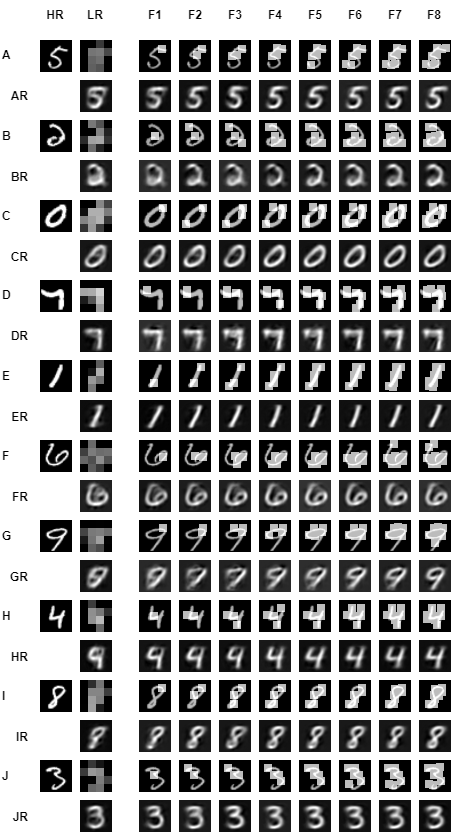}

\caption{Visualization of Fixation Points by FPG2. The first column shows
the original input images. The second column presents the low-resolution
images (odd rows) and the reconstructed images generated by TDFN using
only the low-resolution inputs (even rows). The third through ninth
columns sequentially display the fixation points generated by the
Fixation Point Generator (FPG) (odd rows, represented as light squares)
and the reconstructed images generated by TDFN using both low-resolution
inputs and high-resolution ROIs centered at the fixation points (even
rows). The reconstructed images demonstrate how the inclusion of fixation
points introduces supplementary information, enhancing the reconstruction
quality.}\label{fig:Visualization-of-Fixation}

\end{figure}

\Figref{Visualization-of-Fixation} illustrates the fixation points
generated by FPG2 and their contributions to image reconstruction.
Compared to previous work, the fixation points in this study demonstrate
significantly improved precision, being more accurately located at
stroke endings, corners, intersections, and edges.

\subsection{Comparison of Fixation Number and Classification Accuracy}

\Tabref{Comparison-of-Fixation} compares the recognition performance
of FPG1 and FPG2 under the same number of fixation points. The results
clearly demonstrate that FPG2 significantly enhances classification
accuracy compared to FPG1. For instance, with two fixation points,
classification accuracy increases from 84.2\% to 94.76\%, while with
eight fixation points, it improves from 96.11\% to 99.19\%.

\begin{table}
\centering
\caption{Comparison of Fixation Number and Classification Accuracy}\label{tab:Comparison-of-Fixation}

\begin{tabular}{ccc}
\toprule 
Fixation Point Number & FPG1 Accuracy & FPG2 Accuracy\tabularnewline
\midrule
2 & 84.20\% & 94.76\%\tabularnewline
4 & 92.89\% & 98.07\%\tabularnewline
8 & 96.11\% & 99.19\%\tabularnewline
12 & 97.02\% & 99.38\%\tabularnewline
16 & 97.79\% & 99.39\%\tabularnewline
\bottomrule
\end{tabular}
\end{table}

\subsection{Comparison of Dynamic Fixation Termination}

The fixation procedure can be dynamically terminated when the maximum
classification probability (MCP) output by the classifier exceeds
a predefined threshold, thereby reducing computational cost. This
mechanism is referred to as dynamic fixation termination. \Tabref{Comparison-of-Dynamic}
compares the performance of FPG1 and FPG2 under various MCP thresholds
in terms of classification accuracy, average fixation steps, and coverage
(the proportion of the input image covered by the fixation regions).

\begin{table}
\centering
\caption{Comparison of Dynamic Fixation Termination}\label{tab:Comparison-of-Dynamic}

\begin{tabular}{ccccccc}
\toprule 
\multirow{2}{*}{MCP Threshold} & \multicolumn{3}{c}{FPG1} & \multicolumn{3}{c}{FPG2}\tabularnewline
\cmidrule{2-7}
 & \begin{cellvarwidth}[m]
\centering
Average

Fixation

Steps
\end{cellvarwidth} & Coverage & \begin{cellvarwidth}[m]
\centering
Classification

Accuracy
\end{cellvarwidth} & \begin{cellvarwidth}[m]
\centering
Average

Fixation

Steps
\end{cellvarwidth} & Coverage & \begin{cellvarwidth}[m]
\centering
Classification

Accuracy
\end{cellvarwidth}\tabularnewline
\midrule
0.70 & 0.40 & 2.48\% & 72.32\% & 0.12 & 0.78\% & 89.93\%\tabularnewline
0.75 & 0.56 & 3.50\% & 77.17\% & 0.17 & 1.08\% & 91.58\%\tabularnewline
0.80 & 0.83 & 5.21\% & 81.87\% & 0.23 & 1.45\% & 93.02\%\tabularnewline
0.85 & 1.19 & 7.42\% & 86.62\% & 0.30 & 1.88\% & 94.43\%\tabularnewline
0.90 & 1.72 & 10.76\% & 90.87\% & 0.43 & 2.68\% & 96.27\%\tabularnewline
0.95 & 2.76 & 17.22\% & 95.21\% & 0.64 & 4.00\% & 97.59\%\tabularnewline
0.96 & 3.17 & 19.78\% & 95.89\% & 0.71 & 4.45\% & 97.92\%\tabularnewline
0.97 & 3.69 & 23.04\% & 96.65\% & 0.82 & 5.12\% & 98.31\%\tabularnewline
0.98 & 4.51 & 28.19\% & 97.20\% & 0.97 & 6.05\% & 98.60\%\tabularnewline
0.99 & 6.17 & 38.54\% & 97.53\% & 1.25 & 7.78\% & 98.98\%\tabularnewline
0.993 & -- & -- & -- & 1.41 & 8.78\% & 99.10\%\tabularnewline
0.995 & -- & -- & -- & 1.57 & 9.84\% & 99.29\%\tabularnewline
\bottomrule
\end{tabular}
\end{table}

As shown in \Tabref{Comparison-of-Dynamic}, FPG2 significantly outperforms
FPG1 across all MCP thresholds. With an average of just 1.57 fixation
steps, FPG2 achieves a classification accuracy of 99.29\%, while maintaining
a coverage of only 9.84\%. This highlights the critical role of accurate
fixation point localization in improving classification accuracy and
reducing the average number of fixation steps required.

\section{Conclusion}

This paper proposes a novel approach for generating fixation points
based on the predicted differences between reconstructed images and
input images, replacing the reinforcement learning strategy used in
previous work. This method enables accurate fixation point localization
on high-resolution images, significantly improving the performance
of the TDFN network, including enhanced classification accuracy and
reduced average fixation steps.

It is worth noting that the use of fixation mechanisms, especially
those with precise localization, remains rare in practical visual
tasks. Research in this area is still limited. TDFN is the first implementation
of such a mechanism, and the experimental results demonstrate its
potential. However, further investigation into methods for generating
and utilizing fixation points is necessary.

From a theoretical perspective, the proposed method is not the optimal
solution. Instead, it is tailored to meet the specific requirements
of TDFN, making it more practical to implement within this framework.
We will continue to explore the strengths and limitations of this
approach while seeking improved methods for fixation point generation.

\bibliographystyle{plainnat}
\bibliography{PreciseFixation}

\end{document}